\documentclass{article}

\usepackage[numbers]{natbib}

\usepackage[preprint]{neurips_2024}

\usepackage{mathtools}
\usepackage[utf8]{inputenc} 
\usepackage[T1]{fontenc}    
\usepackage{hyperref}       
\usepackage{url}            
\usepackage{booktabs}       
\usepackage{amsfonts}       
\usepackage{nicefrac}       
\usepackage{microtype}      
\usepackage{xcolor,colortbl}         
\usepackage{xspace}
\usepackage{graphicx}
\usepackage{multirow} 
\usepackage{rotating} 
\usepackage{booktabs} 
\usepackage{tabularx} 
\usepackage{makecell} 
\usepackage{ulem}
\usepackage{caption}
\usepackage{subfigure}
\usepackage{wrapfig}

\usepackage[utf8]{inputenc}
\usepackage{tcolorbox} 

\newcommand{\Tech}{{\textsc{Cond P-Diff}}\xspace}

\newcommand{\bfu}[1]{\textbf{\uline{#1}}}

\title{Conditional LoRA Parameter Generation}

\author{
  Xiaolong Jin$^{1}$\footnotemark[1],\; Kai Wang$^{1}$\thanks{equal contribution, jinxiaolong1129@gmail.com, kai.wang@comp.nus.edu.sg, mtdovent@gmail.com}\; \thanks{corresponding author},\; Dongwen Tang$^{1}$\footnotemark[1],\; Wangbo Zhao$^{1}$,\; \\ \textbf{Yukun Zhou}$^{1}$,\; \textbf{Junshu Tang}$^{2}$,\; \textbf{Yang You}$^{1}$
   \\
  $^{1}$National University of Singapore \quad
  $^{2}$Shanghai Jiao Tong University \\
\vspace{5pt}
Code: \href{https://github.com/NUS-HPC-AI-Lab/Neural-Network-Parameter-Diffusion}{NUS-HPC-AI-Lab/COND P-DIFF}
}

\hypersetup{
    colorlinks=true,
    linkcolor=red,
    citecolor=cyan,
    filecolor=magenta,      
    urlcolor=magenta,
    }

\begin{document}

\maketitle

\begin{abstract}

Generative models have achieved remarkable success in image, video, and text domains. 
Inspired by this, researchers have explored utilizing generative models to generate neural network parameters. 
However, these efforts have been limited by the parameter size and the practicality of generating high-performance parameters. 
In this paper, we propose \Tech, a novel approach that demonstrates the feasibility of controllable high-performance parameter generation, particularly for LoRA (Low-Rank Adaptation) weights, during the fine-tuning process. 
Specifically, we employ an autoencoder to extract efficient latent representations for parameters. 
We then train a conditional latent diffusion model to synthesize high-performing model parameters from random noise based on specific task conditions.
Experimental results in both computer vision and natural language processing domains consistently demonstrate that \Tech can generate high-performance parameters conditioned on the given task. 
Moreover, we observe that the parameter distribution generated by \Tech exhibits differences compared to the distribution obtained through normal optimization methods, indicating a certain level of generalization capability. 
Our work paves the way for further exploration of condition-driven parameter generation, offering a promising direction for task-specific adaptation of neural networks.

\end{abstract}

\section{Introduction}

Recent advancements in generative models  ~\cite{rombachHighResolutionImageSynthesis2022, rameshHierarchicalTextConditionalImage2022, sahariaPhotorealisticTexttoImageDiffusion2022, brown2020language} have marked substantial progress across several domains of artificial intelligence. 
In the computer vision domain, generative adversarial networks ~\cite{goodfellow2014generative}, diffusion models ~\cite{hoDenoisingDiffusionProbabilistic2020}, and other approaches ~\cite{dinh2014nice,rezende2014stochastic} have shown impressive results in image synthesis and manipulation.
Notably, models such as Stable Diffusion ~\cite{rombachHighResolutionImageSynthesis2022}, DALL-E 2 ~\cite{rameshHierarchicalTextConditionalImage2022}, and Imagen ~\cite{sahariaPhotorealisticTexttoImageDiffusion2022} have set new benchmarks in the quality and resolution of generated images.
Moreover, video generation models like Sora ~\cite{OpenAISora2024} have shown promising results in producing coherent and high-quality video sequences, opening new avenues for applications in entertainment and media. 
In the natural language processing domain~\cite{radford2019language,kaplan2020scaling,wei2022chain}, autoregressive models like GPT~\cite{brown2020language} and Llama~\cite{touvron2023llama} have demonstrated promising generation capabilities and alignment with human preference~\cite{jin2024multiverse, ouyang2022training, rafailov2024direct,kadavath2022language}, which underscore the potential of generative models.

Inspired by these achievements, recent studies~\cite{peeblesLearningLearnGenerative2022, wang2024neural} have begun to explore the application of generative models in novel areas, \textit{generating high-performing model parameters}. 
These studies focus on directly generating novel model parameters to accelerate the training process, uncovering parameters that achieve comparable performance with those obtained through conventional optimization methods.

By harnessing the power of generative models, it is possible to substantially reduce the computational cost and time required for model optimization~\cite{peeblesLearningLearnGenerative2022,ruder2016overview,kingma2014adam}. 
Besides, examining the latent relationships between model parameters and performance provides valuable insights into the behavior and characteristics of neural networks~\cite{ha2016hypernetworks}.

However, previous works on parameter generation~\cite{wang2024neural, peeblesLearningLearnGenerative2022, soro2024diffusion, schurholt2022hyper, knyazevParameterPredictionUnseen2021} face several limitations.
On the one hand, the scale of parameters generated by prior methods~\cite{soro2024diffusion,peeblesLearningLearnGenerative2022,wang2024neural} is insufficient for practical applications. For example, G.pt~\cite{peeblesLearningLearnGenerative2022} has been evaluated only on relatively simple datasets such as MNIST and CIFAR-10, which may not sufficiently demonstrate its generalization ability when applied to more complex tasks, and p-diff~\citep{wang2024neural} can generate small-scale high-performance parameters for simple architectures. Besides, ~\cite{schurholt2022hyper} learn a hyper-representation on model zoos for generative use to sample new small-scale model weights.
On the other hand, previous methods do not support conditional high-performance parameter generation. 
P-diff\cite{wang2024neural} lacks support for conditional parameter generation, a crucial feature for real-world applications. 
Although G.pt ~\cite{peeblesLearningLearnGenerative2022} enables controllable parameter generation as an optimizer, it can hardly exhibit comparable performance compared to networks trained by conventional optimization methods. 

Therefore, despite their promising potential, these methods grapple with constraints about parameter size, practicality, and overall performance, which yield the primary question to be explored in this paper:
\textit{(Q) Can we synthesize high-performance parameters conditioned on the given task practically?} 

\begin{wrapfigure}{r}{0.5\textwidth}
  \centering
  \vspace{-15pt}
  \includegraphics[width=0.5\textwidth]{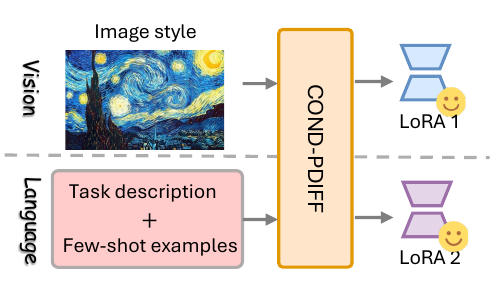}
  \vspace{-20pt}
  \caption{High-performance LoRA parameters generation process by \Tech in vision and language domains.}
  \vspace{-5pt}
  \label{fig:intro-fig}
\end{wrapfigure}

To enhance the practicality of parameter generation, two main challenges exist.
First, parameter generation for complex models entails significant data preparation costs. For example, G.pt~\cite{peeblesLearningLearnGenerative2022} requires training 23 million models, which is infeasible for large models. 
Second, controllable parameter generation is challenging due to the difficulty in modeling the distribution of parameters, making full parameter generation highly complex.
Consequently, we focus on the conditional generation of fine-tuned LoRA (Low-Rank Adaptation) parameters in various domains as LoRA improves downstream task performance with few parameters and a relatively more stable distribution.
Specifically, LoRA ~\cite{hu2021lora} is a parameter-efficient fine-tuning technique that adapts pre-trained models to specific tasks by learning low-rank matrices that modify the model's weights.

To achieve high-performance controllable conditional parameter generation, we propose Conditional Parameter Diffusion, named \Tech, which utilizes a standard latent diffusion model to perform conditional generation, synthesizing a new set of parameters tailored to specific conditions.
Specifically, we use an autoencoder and a conditional latent diffusion model to capture the distribution of network weights. 
First, the autoencoder is trained on a selected set of parameters from models optimized with normal optimization methods, \textit{e.g.}, SGD~\cite{ruder2016overview}, on different datasets, creating latent representations of these parameters. 
Second, we utilize a domain-specific condition, \textit{e.g., text, style image}, projector to encode the condition information and train a conditional diffusion model to reconstruct latent representations.
Finally, as shown in Figure \ref{fig:intro-fig}, the trained conditional latent diffusion model \Tech generates latent representations from random noise in the inference process based on specific task conditions.
Then, the decoder of the trained autoencoder processes these generated representations to produce new, high-performing model parameters.

Our method has the following characteristics: 
i) It demonstrates comparable or superior performance relative to models trained with conventional methods, spanning various datasets and architectures.
ii) The parameters generated by our approach significantly differ from the parameters obtained during normal training, highlighting its capability to synthesize novel parameters rather than merely replicating the training examples.
iii) Extensive evaluations demonstarte the robustness of our approach.
Our method \Tech also shows generalizability in generated high-performance model weights space.
We hope that our findings will provide new insights into the potential of applying conditional diffusion models to parameter generation and highlight a promising direction for task-specific parameter generation of neural networks.

\section{Preliminary}
\subsection{Preliminaries of LoRA}
Low-Rank Adaptation (LoRA) \cite{hu2021lora} enhances the efficiency of fine-tuning large pre-trained language models by minimizing the computational demands usually required for full model retraining. 
LoRA introduces two trainable matrices, $ B \in \mathbb{R}^{d \times r}$ and $A \in \mathbb{R}^{r \times k}$, to each transformer layer. These matrices, where $r$ is much smaller than hidden layer dimension $d$ and task-specific dimension $k$, perform a low-rank approximation of the typical updates made during fine-tuning. 
The core idea is that the necessary adjustments for task-specific adaptation have a low "intrinsic dimension," allowing significant reductions in trainable parameters while maintaining performance.
The pretrained weight matrix $W_{\text{0}}$ remains unchanged, with only $B$ and $A$ being optimized, thus speeding up training and decreasing memory and computational needs. 
The modified forward pass in LoRA is represented as:
\begin{equation}
W_{\text{0}}x + \Delta W x = W_{\text{0}}x + B(Ax)
\end{equation}
where $ \Delta W = BA $ is the update. 
Initially, $B$ is zero, ensuring no changes to $W_{\text{0}}$, and $A$ starts with a small random Gaussian distribution. 
In deployment, the learned low-rank matrices $B$ and $A$ can be integrated into $W_{\text{0}}$.
In this work, we aim to synthesize LoRA parameters because of the practicality and effective LoRA fusion that show the continuous distribution in LoRA parameter space.

\subsection{Preliminaries of Conditional Diffusion Models}
Conditional diffusion models~\cite{hoDenoisingDiffusionProbabilistic2020,rombachHighResolutionImageSynthesis2022,zhang2023adding} extend the standard diffusion model by incorporating conditions into both the forward and reverse processes. 
This conditional information defined by \(c\) allows the model to generate data tailored to specific attributes or requirements.

\textbf{Conditional forward process:} The forward process in conditional models involves adding noise to an initial sample while conditioning on \(c\). 
The probability of transitioning from \(x_{t-1}\) to \(x_t\) under condition \(c\) is modeled as a Gaussian distribution:
\begin{equation}
q(x_t | x_{t-1}, c) = \mathcal{N}(x_t; \sqrt{1 - \beta_t} x_{t-1}, \beta_t \mathbf{I})
\end{equation}
where \(\beta_t\) are the timestep-dependent noise levels, and \(\mathbf{I}\) represents the identity matrix. 
The complete forward process conditioned on \(c\) is given by:
\begin{equation}
q(x_{1:T} | x_0, c) = \prod_{t=1}^T q(x_t | x_{t-1}, c)
\end{equation}

\textbf{Conditional Reverse Process:} 
The reverse process aims to reconstruct the original sample from its noisiest state \(x_T\) conditioned on \(c\). It is formulated by:
\begin{equation}
p_\theta(x_{t-1} | x_t, c) = \mathcal{N}(x_{t-1}; \mu_\theta(x_t, t, c), \Sigma_\theta(x_t, t, c))
\end{equation}
In this process, \(\mu_\theta\) and \(\Sigma_\theta\) are functions estimated by a neural network, which also processes the condition \(c\), ensuring that the recovery of data respects the conditional constraints.

\textbf{Optimization and Inference with Conditions:} 
The training procedure involves minimizing the Kullback-Leibler(KL) divergence between the forward and reverse conditional distributions, specifically:
\begin{equation}
L_{dm} = \mathbb{E}_{q(x_0, c)}\left[ D_{KL}(q(x_{t-1} | x_t, x_0, c) \| p_\theta(x_{t-1} | x_t, c)) \right]
\end{equation}
During inference, the model generates new samples by conditioning on \(c\) and sequentially applying the learned reverse transitions from a noise distribution, enabling the generation of data that closely adheres to the specified conditions.

\section{Methodology}
\label{Methodology}
\subsection{Overview}
\begin{figure*}[ht]
    \centering
    \includegraphics[scale=0.45]{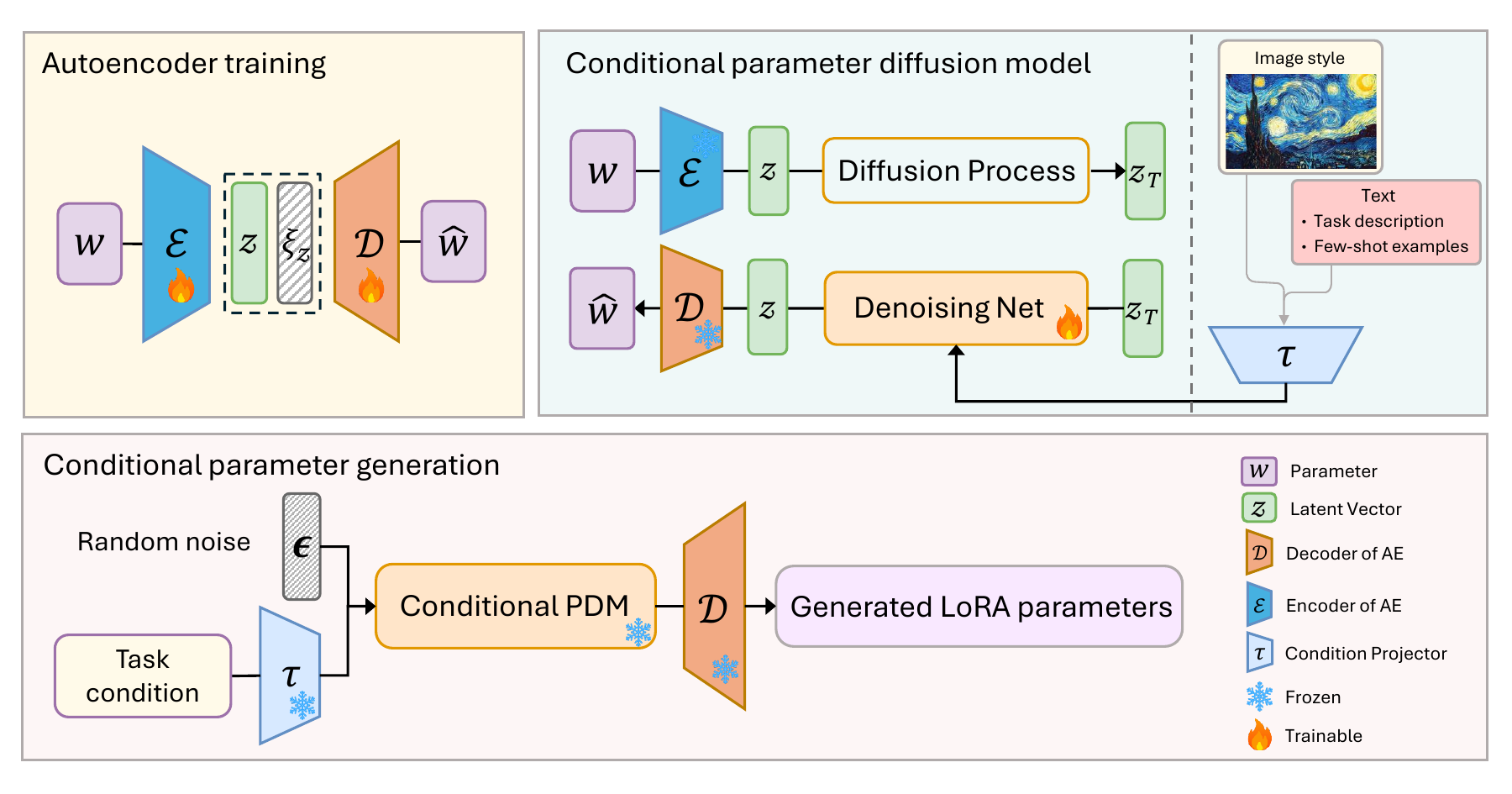}
    \caption{The framework of \Tech. The autoencoder is employed to extract the latent representation of LoRA parameters and reduce memory consumption. The conditional parameter diffusion model aims to synthesize high-performance parameters based on specific task conditions.}
    \label{fig:main-fig-1}
\end{figure*}

We propose conditional parameter generation to synthesize new parameters tailored to specific task conditions. 
Fig \ref{fig:main-fig-1} illustrates our proposed \Tech framework. 
First, given a training dataset of model parameters, we use an autoencoder~\cite{kingma2013auto} to extract latent representations of the parameters and reconstruct the latent vectors by decoder. 
Then, inspired by \cite{wang2024neural}, we train a conditional latent diffusion model to generate high-performance parameters conditioned on specific task information. 
Finally, after training, we employ \Tech by feeding random noise and task-specific conditions into a conditional parameter diffusion model to generate the desired parameters.

\subsection{Parameter autoencoder}
\label{section: parameter AE}
\textbf{Dataset preparation. } 
In this work, we focus on synthesizing LoRA learnable matrix parameters of fine-tuned models by default.
To obtain the training dataset for the parameter autoencoder, we fine-tune the pre-trained model using LoRA on the dataset for task $q$ and collect $N$ different checkpoints in the last $N$ steps.
We denote the training dataset as $\Theta = [\theta_1, \dots, \theta_n, \dots, \theta_N]$, where $\theta_k$ represents the weights of LoRA for the model at a specific fine-tuning stage.
Because the training dataset for \Tech contains model parameters rather than conventional image or language datasets, we propose \textit{task normalization}. Specifically, we employ Z-Score normalization on the parameters of each task individually~\cite{ioffe2015batch}.

\textbf{Training procedure. } Given a training sample $\theta_n$, we flatten parameter matrix $\theta_n$ to a one-dimensional vector $w_n \in \mathbb{R}^{K \times 1}$, which $K$ is the total number of parameter weights of $w_n$.
Then, we utilize an auto-encoder to obtain meaningful and robust latent representations. 
Specifically, we formulate the process as Equation \ref{eq:encode_decode}, where $\mathcal{E}$ and $\mathfrak{D}$ represent the encoder and decoder functions, respectively. 
$z_n$ is the latent representation of the parameter matrix. $\hat{w_n}$ is the reconstruction of parameter $w_n$.
To enhance the generalization and robustness of the autoencoder, we introduce Gaussian noise $\mathbf{\xi_{z}}$ to the latent vector. The final auto-encoder process is formulated as follows:

\begin{subequations}
\begin{equation}
    z_n = \mathcal{E}(w_n) = \text{Encoder}(w_n)
\end{equation}
\begin{equation}
    \hat{w_n} = \mathfrak{D}(z_n) = \text{Decoder}(z_n+\mathbf{\xi_{z}})
\end{equation}
\label{eq:encode_decode}
\end{subequations}

We train the autoencoder function by minimizing loss function below.
\begin{equation}
\mathcal{L} = \frac{1}{N} \sum_{n=1}^N \left\| w_n - \hat{w_n} \right\|^2
\label{eq:loss_function}
\end{equation}

\subsection{Conditional parameter generation}
We utilize a conditional latent diffusion model to synthesize high-performance parameters based on conditions $y$ such as text and image.
To handle different tasks and modalities, we adopt the domain-specific encoder, which is denoted as $\tau_{\text{domain}}(y; \rho)$, where $y$ represents the input condition and $\rho$ denotes the encoder parameters. 
For example, in the NLP experiments of this work, we employ the text decoder in CLIP\cite{radford2021learning}. Inspired by in-context learning, the input condition $y$ consists of a task description and two-shot examples to capture the task information. 
Besides, we utilize stylized images as conditions in style transfer tasks and adopt ResNet~\cite{he2016deep} to extract style latent representations as the condition vector. More details about the condition are shown in Appendix \ref{sec:appendix}.
Regarding the U-Net architecture, we apply one-dimensional convolutions in denoising autoencoders because the weight matrix parameters do not show strong positional relationships different from images where pixels have two-dimensional spatial relationships.

Therefore, given the condition and training parameters samples, we train the conditional latent diffusion model through
\begin{equation}
L_{LDM} := \mathbb{E}_{\epsilon \sim \mathcal{N}(0,1), t} \left[ \|\epsilon - \epsilon_\theta (p_t, t, \tau_{domain, \rho} (y))\|^2 \right], \quad
\label{eq: Conditional parameter generation loss}
\end{equation}
where $\epsilon_\theta$ is learned via Eq. \ref{eq: Conditional parameter generation loss}.
Finally, after conditional diffusion model training, we feed specific conditions corresponding to tasks and random noise to reverse the inference process to obtain high-performing weights for specific tasks.

\section{Experiment}
In this section, we first show the experiment setup. Then, we present the evaluation results, ablation studies, and analysis of \Tech.

\subsection{Experiment setup}
\textbf{Datasets and metrics.} 
We evaluate our method across various domains.
Specifically, in NLP experiments, we test on the language understanding GLUE benchmark \cite{wang2018glue}.
In CV experiments, we focus on the style-transfer tasks. 
We use the SemArt and WikiArt datasets \cite{garciaHowReadPaintings2018, salehLargescaleClassificationFineArt2015}, which contain diverse artistic images, and evaluate them using the Fréchet Inception Distance (FID, \cite{heusel2017gans}, as employed by StyleGAN \cite{karrasStyleBasedGeneratorArchitecture2019}, with lower scores indicating better performance.

\textbf{Dataset collecting and training procedures.} 
In NLP experiments, we collect 150 training samples for models, including BERT, Roberta, GPT-2 by default. 
For instance, in the case of BERT, we fixed pre-trained parameters and fine-tuned the network using LoRA. 
Specifically, we conduct the hyperparameter search for fixed values of $r$ and $\alpha$ and select the fine-tuning hyperparameters that yield the best average performance. 
During the fine-tuning process, we save the checkpoints of the last 150 steps as the training dataset, which includes the LoRA learnable matrix weights. 
In the framework of \Tech, the autoencoder includes 1D CNN-based encoders and decoders. 
We utilize the text encoder from CLIP as the condition text encoder. 
In image style transfer tasks, we fine-tune attention modules of a popular text-to-image model, \textsc{Pix}\textsc{Art}-$\alpha$ model~\cite{chen2024pixartalpha} using LoRA and collected the last 64 LoRA checkpoints of the training process once in 10 steps. 
In the framework of \Tech, we used pre-trained ResNet18 to extract style latent as the condition vector.
All experiments were conducted on the Linux server with four NVIDIA A100 GPUs. The noise $\xi_{z}$ is Gaussian noise with an amplitude of 0.001 by default.
Detailed training hyperparameters for LoRA fine-tuning and \Tech framework are provided in Appendix~\ref{appendix:setup}.

\textbf{Inference procedures.} 
In NLP tasks, we generate 20 LoRA parameters for each task using a conditional diffusion model through random noise and merge these generated parameters into the pre-trained model. 
We select the model that exhibits the best performance on the training dataset and report its performance on the validation dataset.
In style-transfer tasks, we synthesize LoRA parameters of the corresponding styles by feeding the conditional diffusion model with images in various styles as conditions. 
We then merge parameters with \textsc{Pix}\textsc{Art}-$\alpha$'s and utilize them to generate images using a set of prompts.
Finally, we compute the FID score of the generated images.

\textbf{Baselines.} 
1) \textbf{original}: The best validation performance among the originally trained models. 
2) \textbf{model soup}: The validation performance of the model whose weight is the average of the training dataset. Because Mitchell et al. \cite{wortsman2022model} shows averaging the weights of fine-tuned models with different hyperparameter configurations often improves accuracy and robustness.
In style-transfer experiments, we introduce an additional baseline \textbf{no-lora}: we directly employ the predefined \textsc{Pix}\textsc{Art}-$\alpha$ model to demonstrate the effectiveness of LoRA fine-tuning in style-transfer tasks.

\subsection{Experiment results}
\textbf{\Tech can generate high-performance parameters based on task conditions.} 
Table \ref{tab:main-nlp} presents comparison results of \Tech and baseline methods across language understanding GLUE benchmark for three models with different LoRA configurations. 
We observe that \Tech consistently yields comparable performance in most scenarios, demonstrating it learns conditional parameter distributions effectively and stably.
Besides, we note that the baseline \textbf{average}'s performance in some cases surpasses the baseline, validating the potential of model averaging to enhance performance \cite{wortsman2022model}. 

\begin{figure*}[htp]
    \centering
    \subfigure[Visualization of images generated by \Tech parameters in style transfer tasks]
    {\label{fig:cs-exp-visual}\raisebox{0mm}{\includegraphics[width=0.35\textwidth] {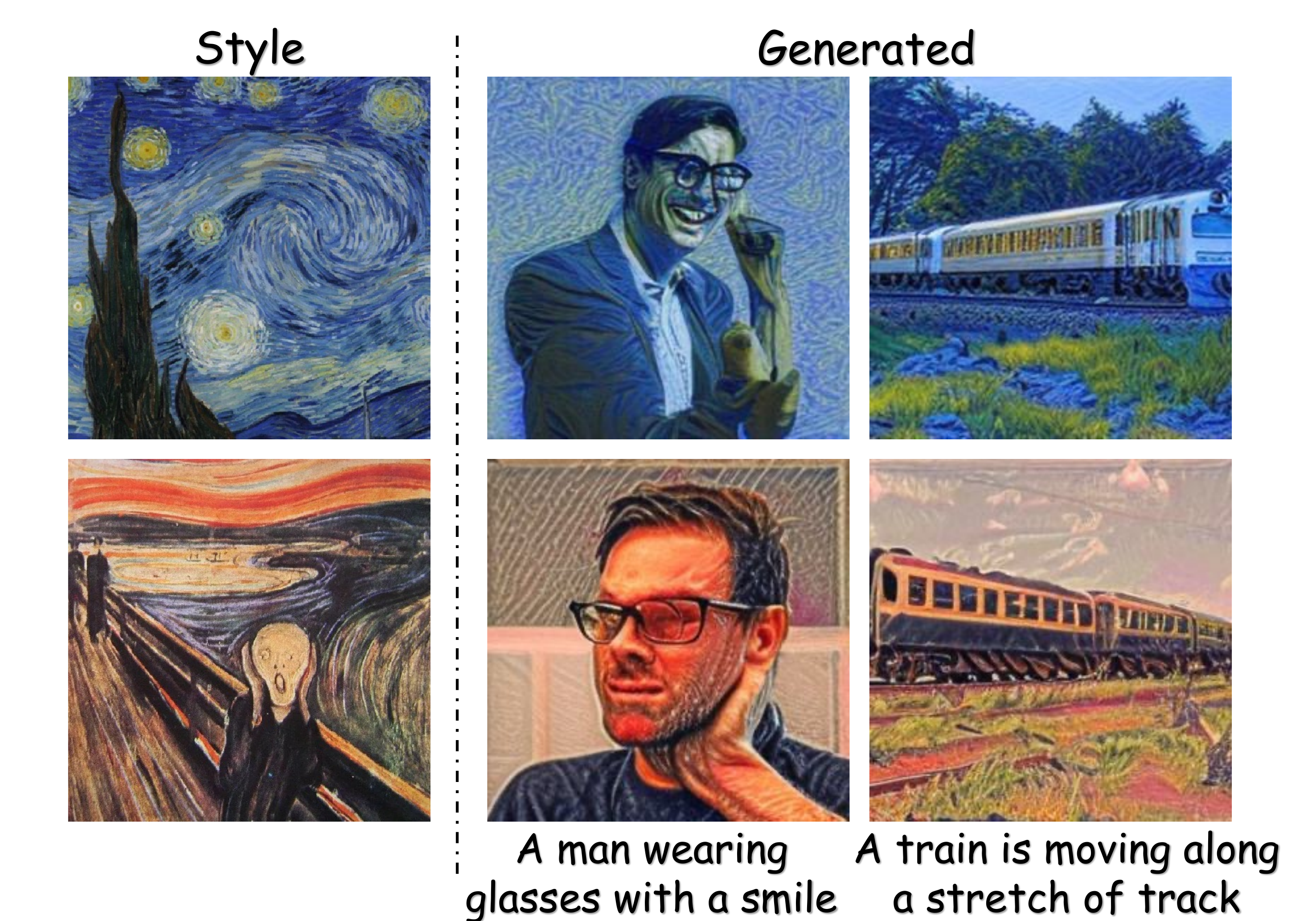}}}\hspace{0.5em}
    \subfigure[t-SNE of the LoRA parameters of original model and generated parameters.]
    {\label{fig:tsne}\raisebox{-0.5mm}{\includegraphics[width=0.25\textwidth]{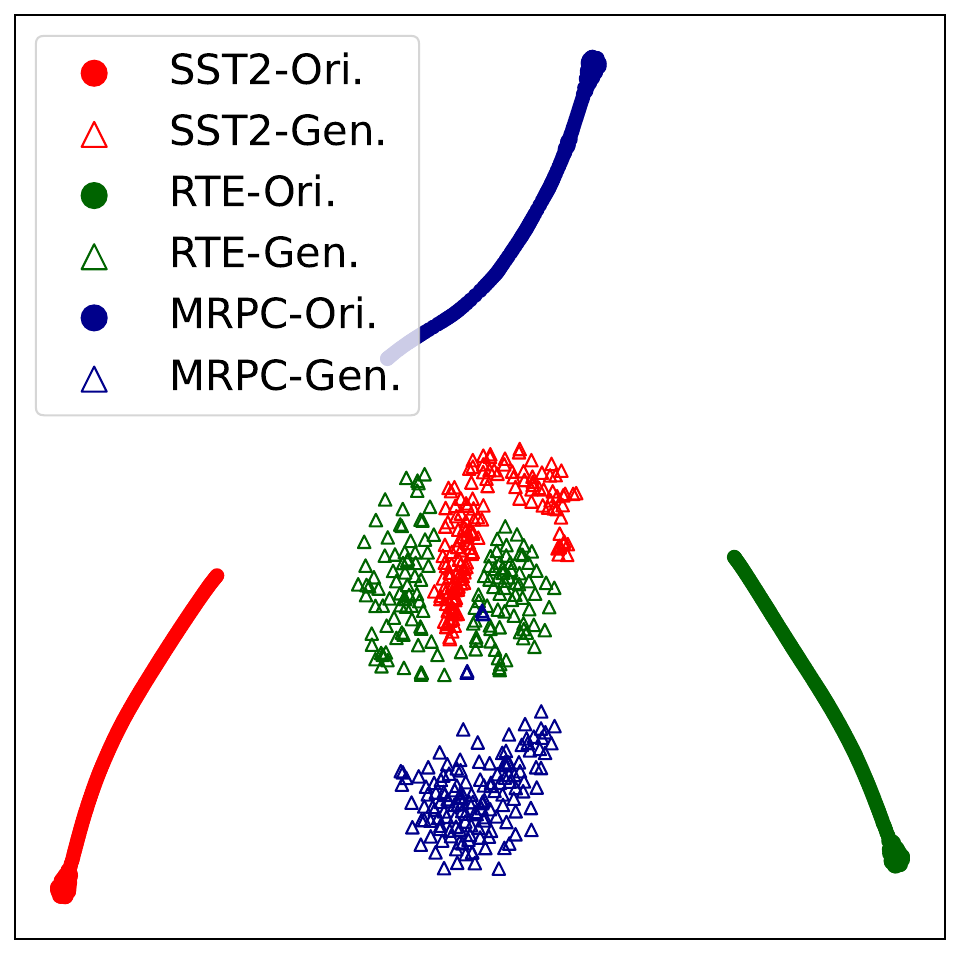}}}
    \subfigure[Similarity comparisons of fine-tuned parameters and parameters generated by \Tech]{\label{fig:similarity}\raisebox{-4mm}{\includegraphics[width=0.32\textwidth]{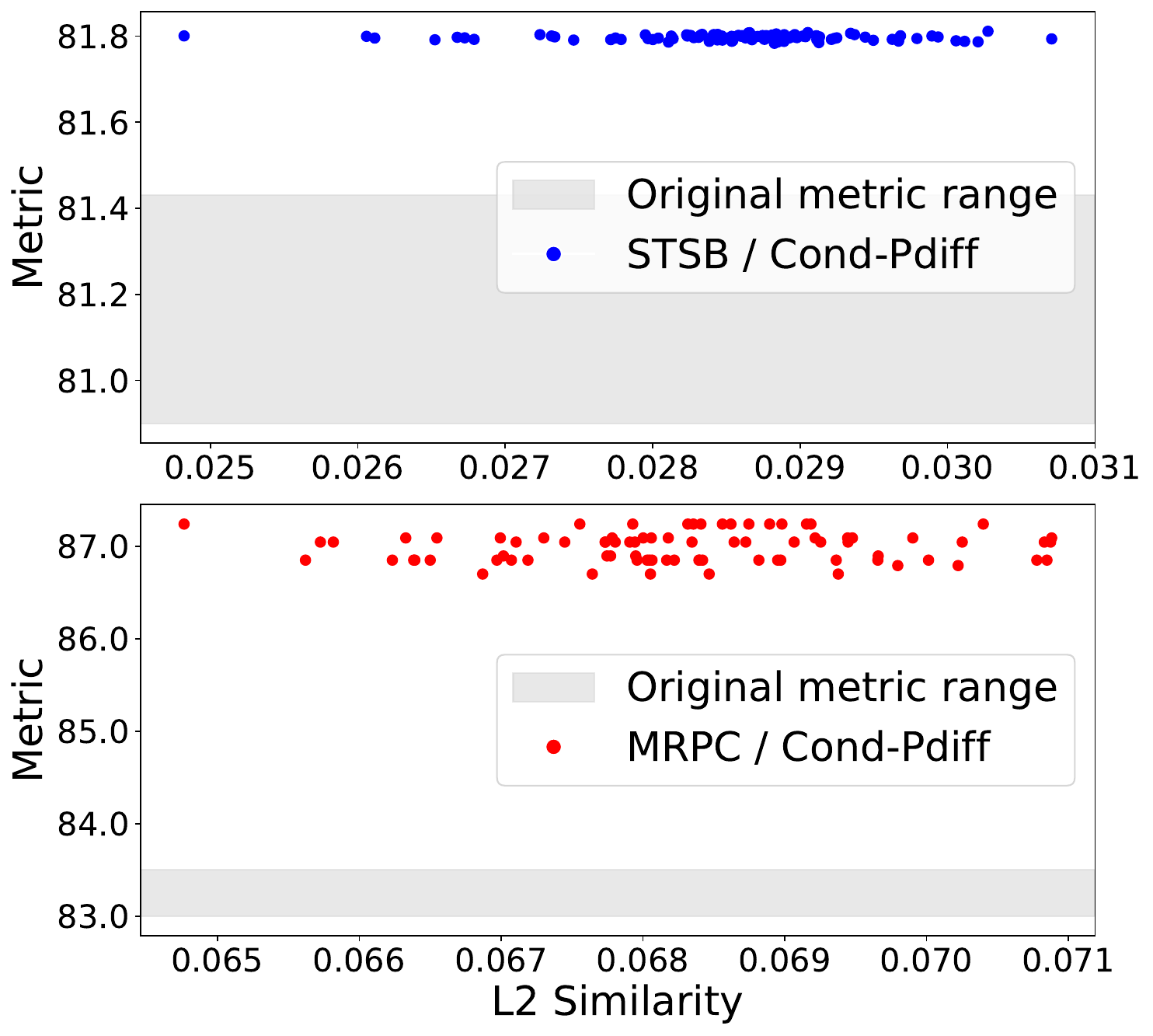}}}
    \vspace{-1em}
    \caption{
    (a) visualize the images generated by \Tech synthetic parameters in style transfer tasks.
    (b) shows the t-SNE of LoRA parameters of the original models, \Tech models on three datasets COLA, QNLI, and STSB. (SST2-Ori. means original parameters and SST-Gen. means generated parameters)
    (c) displays the accuracy and similarity of fine-tuned performance and parameters generated by \Tech. }
    \label{fig:visual-1}
\end{figure*}

Table 2 illustrates the results of \Tech and the baseline in the image style transfer task for different styles. 
We employ the FID \cite{heusel2017gans} to quantitatively assess the quality of style-conditioned image generation. Lower FID represents better image generation quality.
Based on our findings, \Tech efficiently synthesizes specific style-adapted LoRA parameters to generate high-quality images. 
Additional visual results are shown in Figure \ref{fig:cs-exp-visual}.
This demonstrates that \Tech can practically generate high-performance model parameters based on specific conditions.

\begin{table}[htbp]
  \centering
  \tiny
  \setlength{\tabcolsep}{2pt}
\captionsetup{font=small}
  
  \caption{Results of \Tech on GLUE. We present results in the format of '\Tech / orginal / model soup'. \Tech obtains comparable or even better performance than baselines. 'Size' is the parameter size of LoRA. 'Rank' is the parameter $r$ in LoRA. Full' represents fully fine-tuning results.}
    \begin{tabular}{lll|cccccc|c}
    \toprule
    Model & Rank  & \multicolumn{1}{l|}{Size} & \multicolumn{1}{c}{SST2} & \multicolumn{1}{c}{RTE} & \multicolumn{1}{c}{MRPC} & \multicolumn{1}{c}{COLA} & \multicolumn{1}{c}{QNLI} & \multicolumn{1}{c}{STSB} & \multicolumn{1}{c}{Average} \\
    \midrule
    \multirow{5}[3]{*}{BERT} & 1     &   73728    & \bfu{91.6} / 91.6 / 90.8 & 57.4 / \bfu{58.9} / 57.9 & \bfu{87.2 }/ 83.4 / 83.9 & 52.4 / \bfu{52.6} / 52.1 & \bfu{88.7} / 88.7 / 88.1 & \bfu{81.8} / 81.4 / 81.7 & \bfu{76.5} / 76.1 / 75.8 \\
          & 2     &   147456    & \bfu{91.4} / 91.4 / 91.5 & 57.5 / 59.9 / \bfu{60.1} & \bfu{87.3} / 85.1 / 85.5 & \bfu{51.4} / 51.3 / 50.7 & \bfu{88.6 }/ 88.1 / 87.4 & \bfu{82.6} / 81.6 / 81.7 & \bfu{76.5} / 76.2 / 76.2 \\
          & 4     &   294912    & 91.6 / 91.9 / \bfu{92.0} & 62.7 / \bfu{63.2} / 62.8 & 85.4 / 85.4 / \bfu{85.5} & \bfu{53.7} / 53.4 / 52.5 & \bfu{89.8} / 89.6 / 88.9 & 80.6 / \bfu{80.9 }/ 80.7 & 77.3 / \bfu{77.4} / 77.1 \\
          & 16    &    1179648   & \bfu{92.1 }/ 91.6 / 91.5 & 64.2 / 64.3 / \bfu{64.5} & \bfu{87.4} / 87.0 / 86.8 & 56.9 / 57.0 / \bfu{57.5} & 89.8 / 90.1 / \bfu{90.2} & 83.8 / 83.3 / 82.3 & \bfu{79.0} / 78.9 / 78.8 \\
\cmidrule{2-10}          & Full  & 109482240      & 93.5  & 66.4  & 88.9  & 52.1  & 90.5  & 85.8  & 79.5 \\
\midrule
    \multirow{4}[2]{*}{RoBERTa} & 1     &    73728   & 93.3 / \bfu{93.7} / 94.1 & 65.6 / \bfu{68.6} / 68.0 & \bfu{86.9 }/ 84.7 / 85.0 & 49.8 / 50.2 / \bfu{50.5} & \bfu{92.4 }/ 92.0 / 91.4 & 87.3 / 87.5 /\bfu{ 86.9} & 79.2 / \bfu{79.4} / 79.3 \\
          & 2     &    147456   & 93.5 / 93.7 / \bfu{93.8} & 63.2 / 68.2 / \bfu{68.3} & \bfu{87.7} / 85.0 / 84.6 & 50.3 / \bfu{50.7} / 50.6 & \bfu{92.8} / 92.5 / 92.2 & 86.8 / 87.3 /\bfu{ 87.6} & 79.0 / \bfu{79.6} / 79.5 \\
          & 4     &   294912    & \bfu{93.8} / 93.5 / 93.1 & \bfu{69.8 }/ 69.7 / 69.5 & 87.9 / \bfu{88.3} / 87.9 & \bfu{54.1} / 54.0 / 54.1 & 92.0 /\bfu{ 92.4} / 92.9 & 88.3 / 88.2 / \bfu{88.6} & \bfu{81.0} / 81.0 / 81.0 \\
\cmidrule{2-10}          & Full  &  124645632     & 94.8  & 78.7  & 90.2  & 63.6  & 92.8  & 91.2  & 85.2 \\
    \midrule

    \multirow{3}[0]{*}{DeBERTa} & 1     &  92160     & 94.4 / 94.4 / \bfu{94.7} & 61.4 / 61.0 /\bfu{ 61.5} & \bfu{84.0} / 84.0 / 83.2 & 56.8 / \bfu{57.0} / 56.1 & 92.4 / \bfu{92.8} / 92.1 & 87.4 / \bfu{87.8} / 87.0 & 79.4 / \bfu{79.5} / 79.1 \\
          & 2     &  184320     & \bfu{94.9} / 94.8 / 94.0 & \bfu{62.2 }/ 62.1 / 62.0 & \bfu{86.2} / 85.8 / 86.2 & \bfu{58.6} / 58.3 / 57.4 & 92.1 / 92.0 / 92.1 & \bfu{85.2} / 85.2 / 84.5 & \bfu{79.9} / 79.4 / 79.4 \\
          & 4     &    368640   & 94.6 / 94.5 / \bfu{94.7} &\bfu{ 63.2} / 62.8 / 61.9 & \bfu{87.1} / 86.9 / 86.2 &\bfu{ 60.3} / 60.3 / 59.9 & \bfu{93.4} / 93.5 / 93.1 & \bfu{88.7 }/ 88.7 / 88.7 & \bfu{81.2} / 81.1 / 80.7 \\
          \bottomrule
    \end{tabular}%
  \label{tab:main-nlp}%
\end{table}%


\begin{table}[htbp] 
    \centering
    \begin{minipage}{0.45\textwidth}
        \centering
        \captionsetup{font=small}
        
        \caption{FID results of image-transfer tasks. Lower FID is better. Best results are \bfu{bolded}.}
        \scriptsize
    \begin{tabular}{l|cccc}
    \toprule
    Style & original & model soup & no-Lora & \Tech \\
    \midrule
    Van Gogh & \bfu{27.92} & 28.08 & 102.95 & \textbf{28.03} \\
    Edvard & \textbf{27.10} & 27.13 & 96.18 & \bfu{26.98} \\
    Chalk & 36.22 & \bfu{36.00} & 171.82 & 36.18 \\
    Charcoal & 40.80 & \bfu{40.19} & 132.76 & \textbf{40.60} \\
    \midrule
    Average & \bfu{33.01} & 32.86 & 125.93 & \textbf{32.94} \\
    \bottomrule
    \end{tabular}%
        \label{tab:first}
    \end{minipage}
    \hfill 
    \begin{minipage}{0.45\textwidth}
        \centering
        \captionsetup{font=small}
        \caption{Ablation results of training dataset size $N$. Larger $N$ can enhance performances.}
        \scriptsize
        
    \begin{tabular}{l|rrr}
    \toprule
    N     & \multicolumn{1}{c}{SST2} & \multicolumn{1}{c}{STSB} & \multicolumn{1}{c}{MRPC} \\
    \midrule
    1     &     90.23  &   80.71    & 82.71 \\
    100   &   91.63    &  80.91     &83.52 \\
\rowcolor[rgb]{ .918,  .925,  .945}    200   &   91.63&	81.81&	87.24 \\
    500 & 91.63&	81.80&	87.25  \\
    \bottomrule
    \end{tabular}%
        \label{tab:ablation-size}
    \end{minipage}
\end{table}

\begin{table*}[tp]
\caption{Ablation studies of \Tech. We ablate the normalization methods in the training process, the condition representation, and the location of employing \Tech. The Default settings in \Tech are marked in \colorbox[rgb]{0.918, 0.925, 0.945}{gray}. \bfu{Bold entries} are best results.}
\captionsetup{font=scriptsize}
\centering
\tiny
\setlength{\tabcolsep}{3pt}
\subfigure[Comparison among \texttt{no norm.}, \texttt{batch norm.} and \texttt{task norm.}. \texttt{task norm.} can improve performance. ]{
\label{tab:ablation-norm}
\resizebox{0.3\textwidth}{!}{
\begin{tabular}{l|ccc}
    \toprule
    Norm. & SST2 & STSB & MRPC \\
    \midrule
    no norm. &    55.67   &  49.07     & 47.01 \\
    batch norm. &  90.60   &  80.90   & 82.50 \\
        \rowcolor[rgb]{ .918,  .925,  .945} task norm. &  \bfu{91.63}&	\bfu{81.81}&	\bfu{87.24}            \\
    \bottomrule
    \end{tabular}%
    }
}
\hspace{1em}
\subfigure[Few shot examples boost \Tech capability with task information description.]{\label{tab:ablation-condition}
\resizebox{0.32\textwidth}{!}{
    \begin{tabular}{l|rrr}
    \toprule
    Condtion & \multicolumn{1}{c}{SST2} & \multicolumn{1}{c}{STSB} & \multicolumn{1}{c}{MRPC} \\
    \midrule
    one-hot &   90.05  &  77.12 &  80.34 \\
    learnable vector &  90.10     & 80.03      &  81.81\\
    task info &  90.25     & 80.32 & 81.98  \\
        \rowcolor[rgb]{ .918,  .925,  .945} task info+few-shot &  \bfu{91.63}&	\bfu{81.81}&	\bfu{87.24}  \\
    \bottomrule
    \end{tabular}%
}}
\hspace{1em}
\subfigure[\Tech is effective in certain blocks but can boost performance on whole LoRA parameters.]
{
\label{tab:ablation-layers}
\resizebox{0.3\textwidth}{!}
{
    \begin{tabular}{l|rrr}
    \toprule
    LoRA layers & \multicolumn{1}{c}{SST2} & \multicolumn{1}{c}{STSB} & \multicolumn{1}{c}{MRPC} \\
    \midrule
    0-1   &   \bfu{91.63} &	81.43&	83.45      \\
    0-4   &   \bfu{91.63}&	81.45&	83.61      \\
    
    0-8   &  \bfu{91.63}&	81.80   &	85.61      \\
    \rowcolor[rgb]{ .918,  .925,  .945} 0-11  &   \bfu{91.63}&	\bfu{81.81}&	\bfu{87.24}    \\
    \bottomrule
    
    \end{tabular}%
    }
}  
\label{tab:ablations}
\end{table*}

\subsection{Ablation study}
In this section, we conduct multiple ablation studies to report the characteristics of \Tech. We focus on the performance of generated LoRA parameters(rank $r=1$) of BERT on SST2, RTE, and MRPC datasets. The training setting is the same as experiments Table \ref{tab:main-nlp}.

\textbf{Size of the training dataset}
As described in Section \ref{section: parameter AE}, we collect $N$ different checkpoints in the last $N$ steps as a training dataset for task $q$ using LoRA.
We explore the relationship between dataset size $N$ and performance in Table \ref{tab:ablation-size}.  
We observe that the performance improves as the size of the training dataset increases. 
Specifically, a larger training dataset can provide a broader exploration space, thereby enabling \Tech to generate higher performance parameters. 
For instance, performance on the MRPC task improved by 4.53\%.


\textbf{Normalization approach}
As described in Section \ref{section: parameter AE}, we use \textit{task normalization} method.
Table \ref{tab:ablation-norm} shows the impacts of different normalization strategies on performance, including \textit{no norm.}, \textit{batch norm.}, and \textit{task norm.}.
Specifically, \textit{task norm.} refers to normalizing the parameters corresponding to each task individually.
\textit{batch norm.} represents batch normalization.
The experimental setup in Table \ref{tab:ablation-norm} is consistent with that of the experiment in Table \ref{tab:main-nlp}.
We find that \textit{task norm.} consistently yields the best average performance. 
\textit{no norm.} leads to the worst performance because the wide variance in weight distributions across different tasks and outliers 
hinders the convergence of the autoencoder.
Besides, \textit{batch norm.} performed inferior to \textit{task norm.}, as it introduces spurious correlations among parameters across different tasks.

\textbf{Condition information}
The representation of the condition critically affects generation results.
We explore how to represent the task condition effectively to guide conditional parameter generation, as detailed in Table \ref{tab:ablation-condition}. 
Our approach categorizes representations into four types: using one-shot vectors, using only the task description, using only two-shot examples, and using both the task description and two-shot examples. 
Table \ref{tab:ablation-condition} shows that combining the task description with examples yields better outcomes, suggesting that in-context learning can provide more information to establish relationships with the weight parameters.

\textbf{Which part of parameters to synthesis}
We generate LoRA parameters for all blocks by default in Table \ref{tab:main-nlp}. 
To explore the effectiveness of \Tech on different blocks, we present the performance when generating LoRA parameters for only certain blocks.
The experiments in Table \ref{tab:ablation-layers} illustrate that the method is more effective when generating parameters for all blocks. 
We hypothesize that as the number of synthesized parameters increases, the model has a larger exploration space, thereby boosting performance. 
Conversely, performance is constrained by the exploration space and original parameters when focusing on only a subset of parameters.

\begin{figure*}[htp]
    \centering
    \subfigure[Visualization of the interpolation of two generated parameters in different styles.]
    {\label{fig:lora merge}\raisebox{0mm}{\includegraphics[width=0.63\textwidth] {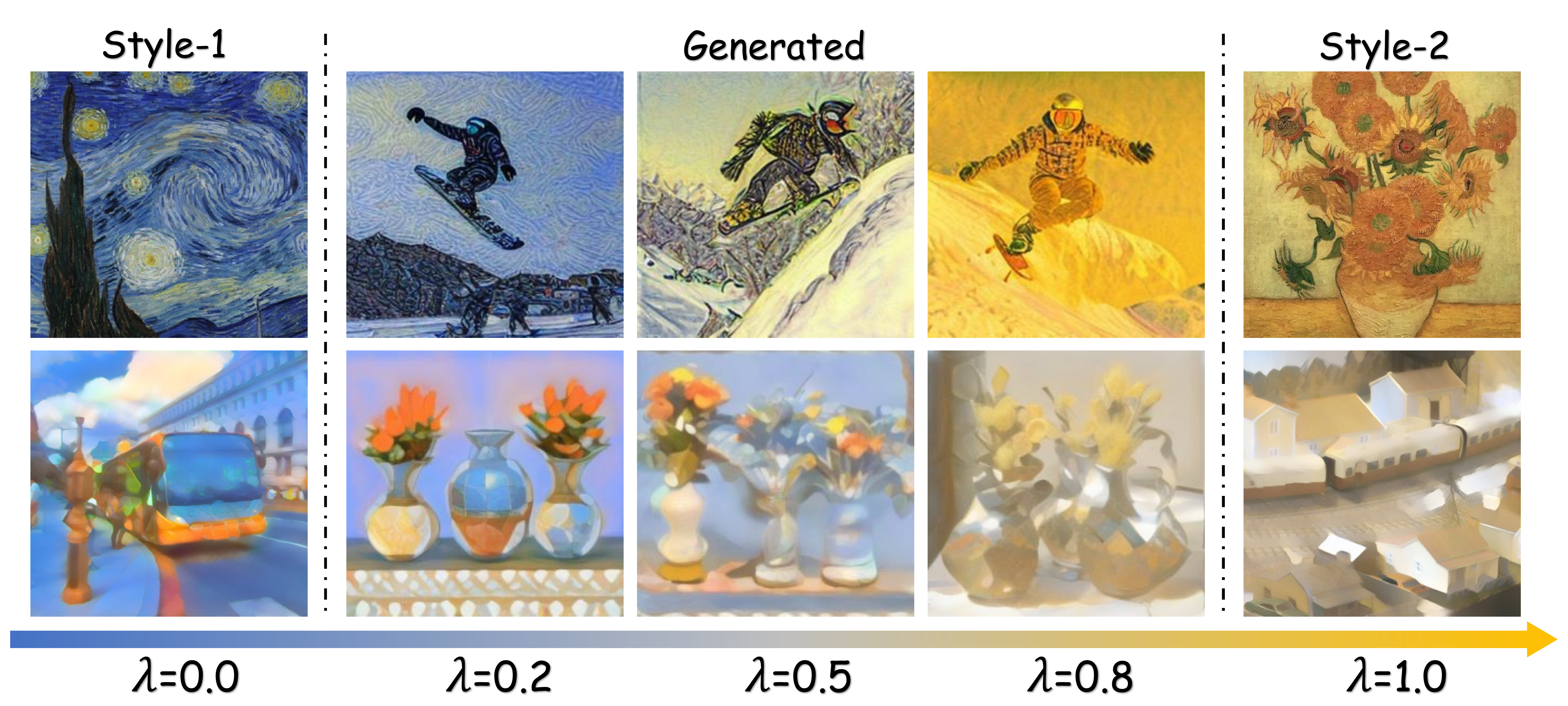}}}\hspace{0.0em}
    \subfigure[Visualization of parameter generation trajectories of \Tech in style-transfer tasks. ]
    {\label{fig:trajectory}\raisebox{0.0mm}{\includegraphics[width=0.28\textwidth]{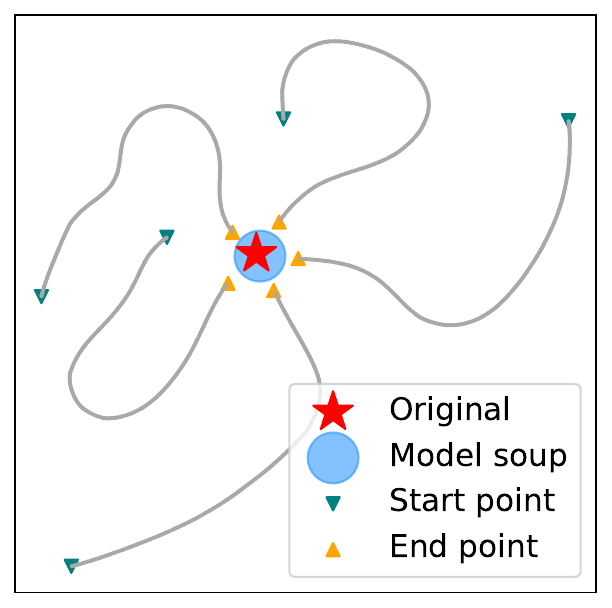}}}\hspace{0.3em}
    \vspace{-1em}
    \caption{
(a) visualizes images generated by interpolated parameters between Style-1 and Style-2. As $\lambda$ increases from left to right, the style gradually shifts towards Style-2 from Style-1.
(b) exhibits the generated parameters' trajectory at different
time steps during the inference stage using t-SNE from five random noise start points in image-transfer tasks.}
    \label{fig:diversity}
\end{figure*}

\subsection{Analysis}
In this section, we conduct a detailed analysis of \Tech.
Specifically, we explore two critical questions: First, does \Tech merely replicate training data, or can it generate high-performance model parameters that are distinct from the originals? 
Second, does the generated parameter space of \Tech have generalizability?

\textbf{\Tech is not merely cloning model parameters.} 

\textbf{Similarity vs. Performance }
First, we calculate the $L_2$ distance between the generated and original parameters. Figure~\ref{fig:similarity} illustrates the relationship between the similarity of the generated parameters and performance.
We observe that \Tech attains various similarities and achieves better performance compared to original fine-tuned weights across various datasets.

\textbf{Parameter distribution}
We employ t-SNE~\cite{van2008visualizing} to analyze the distributions of generated parameters and original weights of fine-tuned models on datasets COLA, QNLI, and STSB, as shown in Figures \ref{fig:tsne}.
We observe that the distribution of generated parameters by \Tech significantly differs from the original parameters. 
The distribution of the original parameters can be viewed as following the trajectory of the optimization process. 
In contrast, \Tech generates novel high-performance parameters by learning the distribution of parameters.
Besides, the high-performance parameters generated by \Tech are dispersed more broadly, underscoring the generative model’s potential to identify novel high-performance parameters beyond traditional optimization pathways.
Interestingly, the high-performance parameter distributions generated by \Tech for the three datasets are very similar, demonstrating the necessity of exploring the high-performance parameter space.

\textbf{Trajectories of \Tech process.}
Figure \ref{fig:trajectory} visualizes the generated parameters at different time steps during the inference stage using t-SNE~\cite{van2008visualizing} to explore the generation process in the image style-transfer tasks.
We display five trajectories initialized from five different random noises and present the model soup and the original model parameters.
The parameters derived from the model soup are located near the original parameters. 
We observe that the generated parameters gradually approach the original parameters but ultimately maintain some distance from them, indicating that \Tech generates high-performance parameters that are distributed differently from the original parameters rather than directly replicating them.
The variations in the trajectories also demonstrate the robustness of \Tech.

\textbf{Generalizability}
We examine the generalization of the generated parameter space in the task of image style transfer. 
We select parameters, $\theta_{\text{style1}}$ and $\theta_{\text{style2}}$, generated by \Tech conditioned two distinct styles, style1 and style2. 
To interpolate between these styles, we compute a new set of parameters $\theta_{\text{interp}}$ as $\theta_{\text{interp}} = (1 - \lambda) \theta_{\text{style1}} + \lambda \theta_{\text{style2}}$, where $\lambda \in [0,1]$ is the interpolation factor.
Subsequently, we evaluate the effectiveness of $\theta_{\text{interp}}$ in style transfer.
Figure \ref{fig:trajectory} illustrates the visualization of images generated by interpolated parameters between Style-1 and Style-2. 
As $\lambda$ increases from left to right, the style gradually shifts towards Style-2. 
The continuous style change demonstrates the generalization of the generated parameter space. We also explore the generalization of the condition space in the Appendix \ref{appendix:generalizability}

\section{Related work}
\noindent\textbf{Diffusion models} 
Diffusion models~\cite{hoDenoisingDiffusionProbabilistic2020, dhariwalDiffusionModelsBeat2021, peeblesScalableDiffusionModels2023} have recently emerged as a powerful class of generative models, enabling high-fidelity synthesis of complex data distributions.
The research on the diffusion model can be generally classified into four categories. 
The first category aims to enhance image synthesis quality \cite{rombachHighResolutionImageSynthesis2022,rameshHierarchicalTextConditionalImage2022,sahariaPhotorealisticTexttoImageDiffusion2022} 
Second, researchers focus on accelerating the sampling process \cite{songDenoisingDiffusionImplicit2022, luDPMSolverFastODE2022}. 
Third, recent research has also focused on reevaluating diffusion models through the lens of continuous analysis like score-based generative modeling \cite{fengScoreBasedDiffusionModels2023}.
Fourth, the success of diffusion models has sparked their application in various domains, \cite{kongDiffWaveVersatileDiffusion2021, luoDiffusionProbabilisticModels2021,wollebDiffusionModelsMedical2022}. In this work, we explore the conditional diffusion model in the parameter generation domain.

\noindent\textbf{Conditional generation}
Conditional generation has gained significant attention in computer vision and natural language processing. 
Three prominent frameworks have emerged: conditional GANs \cite{mirzaConditionalGenerativeAdversarial2014, isolaImagetoImageTranslationConditional2018,zhuUnpairedImagetoImageTranslation2020}, conditional VAEs \cite{sohnLearningStructuredOutput2015, yanAttribute2ImageConditionalImage2016}, and conditional diffusion models xw\cite{rombachHighResolutionImageSynthesis2022, hoDenoisingDiffusionProbabilistic2020}, which incorporate conditions to guide the generation process, enabling the creation of visually coherent and semantically meaningful data samples.
Conditional GANs incorporate condition information into GAN to generate images conditioned on specific attributes or labels.
Conditional diffusion models take this further by generating visually coherent and semantically meaningful images from the textual description, demonstrating superior image synthesis quality compared to GANs. 
Building upon the success of conditional diffusion models, we propose to extend this approach to generating neural network parameters based on specific conditions.

\noindent\textbf{Parameter generation}
The field of parameter generation has seen significant progress in recent years, with HyperNetworks (\citep{ha2016hypernetworks} and generative models of neural network checkpoints \cite{peeblesLearningLearnGenerative2022} emerging as promising approaches. 
\cite{ha2016hypernetworks} introduced HyperNetworks, which uses a hypernetwork to learn the parameters for another neural network.
\cite{finnModelAgnosticMetaLearningFast2017} proposes Model-Agnostic Meta-Learning, which learns an initialization for efficient fine-tuning.
\cite{peeblesLearningLearnGenerative2022} introduce the model G.pt to predict the distribution over parameter updates given an initial input parameter vector and a prompted loss or error.
\cite{schurholt2022hyper} trained autoencoder on a model zoo to learn a hyper-representation for generative use to sample new model weights
\cite{knyazevParameterPredictionUnseen2021} use a GNN-based model to sample network parameters.
\cite{erkocc2023hyperdiffusion} directly leverages MLP weights and generates neural implicit fields encoded by synthesized MLP weights.
\cite{wang2024neural} uses a diffusion model to generate high-performing neural network parameters across various architectures and datasets.
Different from the previous works, we focus on conditional parameter generation to generate high-performing weights based on specific task conditions practically.

\section{Conclusion}
In this work, we proposed an approach \Tech for high-performance controllable parameter generation, specially for LoRA parameters. 
We utilize an autoencoder and a conditional latent diffusion model to capture the distribution of high-performing parameters and perform conditional generation, synthesizing a new set of parameters tailored to specific conditions. 
We show that our method can efficiently synthesize novel and high-quality model parameters.
The parameter distribution generated by \Tech exhibits differences compared to the distribution obtained through conventional optimization methods, indicating a certain level of generalization capability.

\subsection{Limitation and future work}
Nonetheless, it is essential to recognize that diffusion in parameter generation is still largely unexplored despite the significant advances in the realm of image and video synthesis.
In this work, we present a preliminary methodology for conditional parameter diffusion. 
However, several challenges remain unresolved, including reducing memory demands for large model architectures, enhancing the generalizability of generation techniques, and improving the representation of dataset conditions.
Furthermore, integrating knowledge graphs with conditional diffusion offers promising directions for controlling conditional generation.

\bibliographystyle{plain}
\bibliography{references}

\newpage
\appendix
\label{sec:appendix}

\section{Detailed related work}

\noindent\textbf{Diffusion models} 
Diffusion models have emerged as a powerful class of generative models, enabling high-fidelity synthesis of complex data distributions.
Diffusion models are based on non-equilibrium thermodynamics, which gradually add noise to data and learn to reverse the diffusion process to generate samples.~\cite{hoDenoisingDiffusionProbabilistic2020, dhariwalDiffusionModelsBeat2021, peeblesScalableDiffusionModels2023}
The research on the diffusion model can be generally classified into four categories. 
The first category aims to enhance image synthesis quality, as demonstrated by notable models such as Stable Diffusion \cite{rombachHighResolutionImageSynthesis2022}, DALL·E 2 \cite{rameshHierarchicalTextConditionalImage2022}, and Imagen \cite{sahariaPhotorealisticTexttoImageDiffusion2022} by leveraging techniques like CLIP-based text encoders, latent space diffusion, and hierarchical architectures. 
Second, researchers focus on accelerating the sampling process, with key developments including Denoising Diffusion Implicit Models \cite{songDenoisingDiffusionImplicit2022} and DPM-Solver \cite{luDPMSolverFastODE2022}. These approaches aim to improve the computational efficiency of diffusion models through deterministic sampling, closed-form expressions, and numerical ODE solvers.
Third, recent research has also focused on reevaluating diffusion models through the lens of continuous analysis like score-based generative modeling \cite{fengScoreBasedDiffusionModels2023} in continuous-time settings.
Fourth, the success of diffusion models has sparked their application in various domains, including text-to-speech synthesis \cite{kongDiffWaveVersatileDiffusion2021}, 3D shape generation \cite{luoDiffusionProbabilisticModels2021}, and anomaly detection in medical images \cite{wollebDiffusionModelsMedical2022}, demonstrating the potential of diffusion models beyond image synthesis. In this work, we explore the conditional diffusion model in the parameter generation domain.

\noindent\textbf{Conditional generation}
Conditional generation has gained significant attention in machine learning, particularly in computer vision and natural language processing. 
Three prominent frameworks have emerged: conditional GANs \cite{mirzaConditionalGenerativeAdversarial2014, isolaImagetoImageTranslationConditional2018,zhuUnpairedImagetoImageTranslation2020}, conditional VAEs \cite{sohnLearningStructuredOutput2015, yanAttribute2ImageConditionalImage2016}, and conditional diffusion models \cite{rombachHighResolutionImageSynthesis2022, hoDenoisingDiffusionProbabilistic2020}, which incorporate conditions to guide the generation process, enabling the creation of visually coherent and semantically meaningful data samples.
Conditional GANs incorporate condition information into GAN to generate images conditioned on specific attributes or labels.
Conditional diffusion models take this further by generating visually coherent and semantically meaningful images from the textual description, demonstrating superior image synthesis quality compared to GANs. 
Building upon the success of conditional diffusion models, we propose to extend this approach to generating neural network parameters based on specific conditions. 

\noindent\textbf{Parameter generation}
The field of parameter generation has seen significant progress in recent years, with HyperNetworks (\citep{ha2016hypernetworks} and generative models of neural network checkpoints \cite{peeblesLearningLearnGenerative2022} emerging as promising approaches. 
\cite{ha2016hypernetworks} introduced HyperNetworks, which uses a hypernetwork to learn the parameters for another neural network.
\cite{finnModelAgnosticMetaLearningFast2017} proposes Model-Agnostic Meta-Learning, which learns an initialization for efficient fine-tuning.
\cite{peeblesLearningLearnGenerative2022} introduce the model G.pt to predict the distribution over parameter updates given an initial input parameter vector and a prompted loss or error.
\cite{schurholt2022hyper} trained autoencoder on a model zoo to learn a hyper-representation for generative use to sample new model weights
\cite{knyazevParameterPredictionUnseen2021} use a GNN-based model to sample network parameters.
\cite{erkocc2023hyperdiffusion} directly leverages MLP weights and generates neural implicit fields encoded by synthesized MLP weights.
\cite{wang2024neural} uses a diffusion model to generate high-performing neural network parameters across various architectures and datasets.
Different from the previous works, we focus on conditional parameter generation to generate high-performing weights based on specific task conditions practically.

\section{Experiment setup}
\label{appendix:setup}
In this section, we show detailed experiment setups, including dataset information and training configuration.

\subsection{Style transfer experiments}
In this section, we provide detailed information about the training configurations used for both the autoencoder and the diffusion model in the style transfer task.

\textbf{Autoencoder configuration:} 
The encoder is a 1D CNN-based model where the channel of each layer is $(16, 32, 64, 128, 256, 384, 512, 768, 1024, 64)$. 
At the bottom layer, we flatten the parameters and map them to a latent dimension of $256$ with a linear layer. 
In the decoder part, we use transposed convolutions with the same number of channels and layers to upsample back to the original shape. 

The training details of hyperparameters are as follows: total number of parameters $516,096$, kernel size for CNN model $9$, learning rate $2\times10^{-4}$ with cosine annealing, total training steps $12,000$, batch size $64$. In addition, to reduce memory usage and accelerate computations, mixed-precision is enabled with $bfloat16$ for the first $75\%$ of the training process.

\textbf{Diffusion Model configuration:}
The architecture of the DDPM comprises a 1D CNN-based U-Net \cite{UNetImageSegmentation} with channels $(64, 128, 256, 512, 768, 1024, 1024, 32)$. A fully connected layer is applied at the bottom of the U-Net after flattening. 
In addition to the U-Net, we employ a style feature extraction network as the condition projector, consisting of two convolutional layers, an average pooling layer, and a fully connected layer.
The extracted features are added as embeddings to the bottom layer of the U-Net.
The training details of hyperparameters are as follows: kernel size for CNN model $3$, learning rate $5\times10^{-4}$ with cosine annealing, total training steps $50,000$, batch size $128$, number of diffusion steps $1,000$, $\beta$ in the diffusion model shifted linearly from $0.0001$ to $0.02$ in diffusion models. And the same as AE training, mixed-precision is enabled with $bfloat16$ for the first $75\%$ of the training process.

\begin{figure*}[ht]
    \centering
    \includegraphics[scale=0.57]{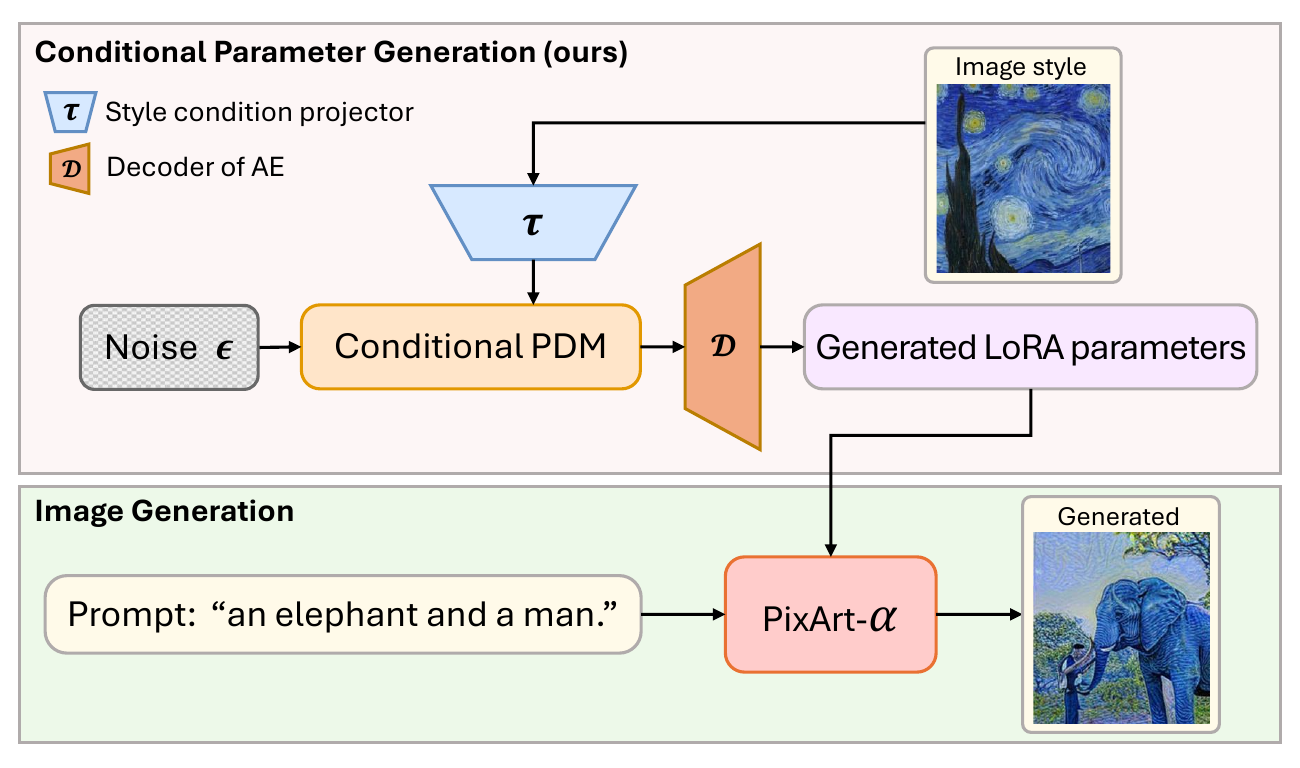}
    \caption{\Tech framework in style-transfer tasks.}
    \label{fig:appendix-cv-fig-1}
\end{figure*}

\textbf{Framework:}
This section describes the framework and workflow of the style transfer task with our conditional parameter generation in detail, as illustrated in Figure \ref{fig:appendix-cv-fig-1}.

\textbf{Data Preparation:} The first step is selecting appropriate data, including style image and parameter data. 
For style image data, we select a total of 16 groups of data with different styles. 

7 groups, such as \textit{Van Gogh}, \textit{Edvard}, and \textit{Jacoulet}, are manually selected from \textit{SemArt} and \textit{WikiArt} \cite{garciaHowReadPaintings2018, salehLargescaleClassificationFineArt2015} datasets, which totally includes more than 250,000 works by 3,000 artists. 
The other 9 groups, such as \textit{Chalk} and \textit{Charcoal}, are generated by a traditional image style transfer algorithm \cite{gatysImageStyleTransfer2016} to make sure the styles of images in a particular group are highly consistent. 
For parameter data, we use the \textit{PixArt-$\alpha$} \cite{chen2024pixartalpha} as the base model, which is a transformer-based text-to-image diffusion model with smaller parameter sizes and competitive quality. We finetuned it with the style image data.
Each set of LoRA parameters holds 64 checkpoints from the last 64 steps of one training. 
Thus, we obtained 16 sets of parameter data, with 64 LoRA parameters in each set.

\textbf{Training of Autoencoder and Conditional Parameter Diffusion:} 
We introduce details of the training process of the autoencoder and the diffusion models.
For the autoencoder, we use the parameter data to train the autoencoder to encode the LoRA parameters into a 256-dimensional latent space. Note that we did not use the style image data in this process. 
For conditional diffusion model, we use style condition extractor to extract the style features of the style image data, and merge the features into the diffusion model as condition information.

\textbf{Generation Process:} 
The generation process is divided into two steps. 
First, the LoRA parameters are obtained by the conditional parameter diffusion model, and then they are merged into PixArt-$\alpha$ to obtain the style image. 

\textbf{Parameter Generation:} 
In the inference process, the diffusion model is fed with noise and an image in a particular style as conditions, and the generated latent is fed into the decoder to get completed LoRA parameters. 

\textbf{Image Generation:} Next, merge the generated LoRA parameters to PixArt-$\alpha$. 
Then, we get the PixArt-$\alpha$ finetuned with a particular style. 
Then, we can feed it with a prompt to get an image whose style corresponds to our input condition.

\subsection{Language experiments}
\subsubsection{Datasets}
In NLP tasks, we use \textbf{GLUE benchmark} \cite{wang2018glue}, a benchmark for evaluating natural language understanding capabilities. \textbf{SST2} \cite{SST2socher-etal-2013-recursive}: A sentiment analysis benchmark using movie review excerpts, labeled as positive or negative, to aid in sentiment understanding. \textbf{RTE}: A dataset for evaluating if one sentence logically entails another, testing models' understanding of textual entailment. \textbf{MRPC} \cite{MRPCbadsha2018mrpc}: Contains sentence pairs to benchmark models' paraphrasing and semantic equivalence capabilities. \textbf{CoLA}: Tests language models' grasp of English grammar, with sentences labeled as grammatically acceptable or not. \textbf{QNLI}: Converts question-answer pairs into inference tasks, assessing if sentences are correct responses to questions. \textbf{STSB} \cite{STSBcer2017sts}: A benchmark for measuring semantic similarity between sentences, rated on a scale from 0 to 5 for nuanced meaning comprehension.

\subsubsection{LoRA configurations}

In this section, we introduce the configuration of LoRA fine-tuning as presented in Table \ref{tab:main-nlp}.
All models are fine-tuned with 20 epochs and a dropout rate of 0.1. 
Mixed-precision training is enabled with FP16 to accelerate computation and reduce memory usage.
The learning rate is set to 0.0001, and a warmup ratio of 0.1 is used to gradually increase it at the beginning of the training. 
Additionally, a weight decay of 0.1 is applied to regularize the model and prevent overfitting.

\begin{table}[htbp]
  \centering
  \caption{Add caption}
    \begin{tabular}{c|cccc|cccc|ccc}
    \toprule
    Model & \multicolumn{4}{c|}{BERT}     & \multicolumn{4}{c|}{RoBERTa}  & \multicolumn{3}{c}{DeBERTa} \\
    \midrule
    Rank  & 1     & 2     & 4     & 16    & 1     & 2     & 4     & 16    & 1     & 2     & 4 \\
    \midrule
    alpha & 8     & 8     & 16    & 32    & 8     & 8     & 16    & 32    & 8     & 8     & 16 \\
    \bottomrule
    \end{tabular}%
  \label{tab:addlabel}%
\end{table}%

\subsubsection{Condition}
\begin{tcolorbox}
This is task 'SST-2'. SST-2 (The Stanford Sentiment Treebank) includes sentences from movie reviews and their sentiment labels (positive or negative). It tests a model's ability to capture sentiment from text.
\\

Example 1: Sentence: "The movie was fantastic!" Label: Positive. 
Example 2: Sentence: "I did not enjoy the film at all." Label: Negative.
\end{tcolorbox}

\begin{tcolorbox}
This is task 'RTE.' RTE (Recognizing Textual Entailment) involves pairs of sentences and asks whether the second sentence is true (entails), false, or undetermined based on the information in the first sentence.
\\

Example 1: Sentence 1: "The cat sat on the mat." Sentence 2: "There is a cat on the mat." Label: Entailment.
Example 2: Sentence 1: "Sarah bought two tickets to Hawaii for her honeymoon." Sentence 2: "Sarah is planning a trip to Hawaii." Label: Entailment.
\end{tcolorbox}

\begin{tcolorbox}
This is task 'MRPC'. MRPC('Microsoft Research Paraphrase Corpus') checks if sentences are paraphrased from each other.
\\

Example 1: "The storm left a wake of destruction." / "Destruction was left by the storm." -> Paraphrase.
Example 2: "He says that he saw the man leave." / "He says the man stayed in." -> Not Paraphrase.''',
\end{tcolorbox}

\begin{tcolorbox}
This is task 'COLA'. CoLA (The Corpus of Linguistic Acceptability) consists of English sentences labeled as grammatically correct or incorrect. It's designed to evaluate a model's ability to understand English grammar.
\\

Example 1 : Sentence: "The cat sat on the mat." Label: Correct. Sentence: "On the mat sat cat." Label: Incorrect.

Example 2: Sentence: "She reads books every day." Label: Correct. Sentence: "Books every day reads she." Label: Incorrect.
\end{tcolorbox}

\begin{tcolorbox}
This is task 'QNLI'. QNLI (Question Natural Language Inference) involves pairs of a question and a sentence, where the goal is to determine whether the sentence contains the answer to the question.
\\

Example 1: Question: "What color is the sky?" Sentence: "The sky is usually blue." Label: Entailment.
Example 2: Question: "Who wrote '1984'?" Sentence: "George Orwell is the author of 'Animal Farm' and '1984'." Label: Entailment.
\end{tcolorbox}

\begin{tcolorbox}
This is task STSB. STSB(Semantic Textual Similarity Benchmark) aims to rate sentence pair similarity on a 0-5 scale.
\\

Example 1: "A man is playing a guitar." / "A man is playing an instrument." -> Score: 4.5.
Example 2: "A child is riding a horse." / "A horse is being ridden by a child." -> Score: 5.
\end{tcolorbox}

\begin{figure*}[ht]
    \centering
    \includegraphics[scale=0.132]{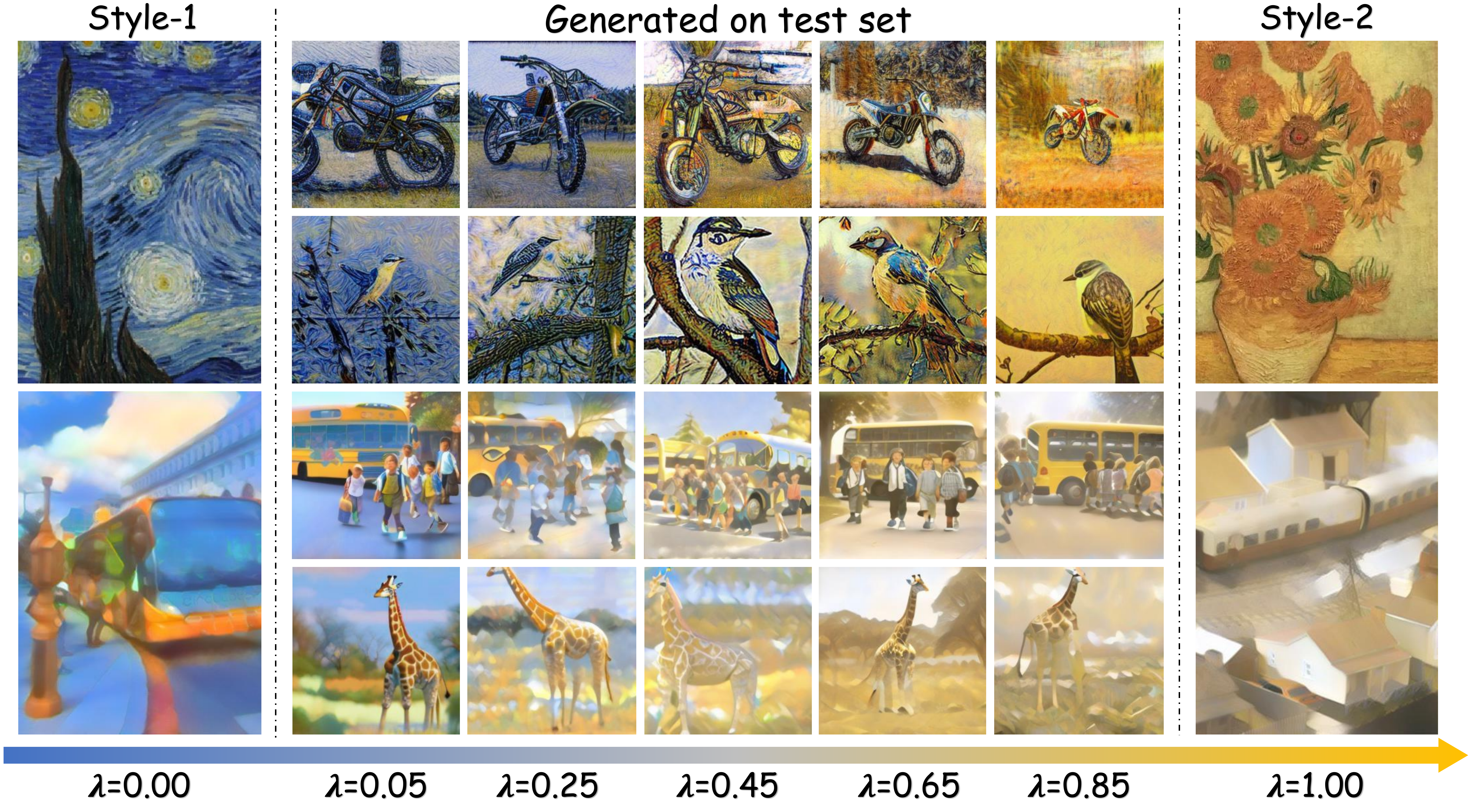}
    \caption{Visualization of the image generated by LoRA parameters, which is generated by \Tech on the test set with conditions that the model has never seen.}
    \label{fig:appendix-cv-fig-2}
\end{figure*}

\section{Explorations of \Tech generalizability}
\label{appendix:generalizability}
We consider that the generalizability of \Tech is limited by the current amount of data. If we want the model to gain generalizability, we need to sample enough LoRA parameters in the parameter space, which is difficult to achieve. Therefore, in this experiment, we first make a style-continuous dataset, which can be equivalent to sampling enough data points in a subspace to provide enough data for our model. We then trained our model on the style-continuous dataset we created to verify its generalizability.

\textbf{Make a style-continuous dataset:}

Since it is difficult to find style-continuous data, we use some AI-generated images to make a style-continuous parameter-image pair dataset, to verify the continuity of the parameter space and the model's generalization ability. Here are the detailed steps:

Firstly, we train the LoRA parameters relevant to style-1 using style-1 images; train the parameters relevant to style-2 using style-2 images. Next, we use formula $\theta_{\text{interp}} = (1 - \lambda) \theta_{\text{style1}} + \lambda \theta_{\text{style2}}$ to combine LoRA parameters in different proportions to obtain 1000 LoRA parameters between style-1 and style-2 ($\lambda$ is from $\{0.000, 0.001, 0.002,\cdots ,0.999\}$). Then we merge the 1000 LoRA parameters to PixArt-$\alpha$ in turn and randomly select some prompts to generate images in relevant style. Thus, we obtain a dataset of 1000 parameter-image pairs.

\textbf{Train on the style-continuous dataset:}

With the above style-continuous parameter-image pair data, we can verify continuity of the parameter space and the generalization ability of our model. The detailed training process is as follows:

First, we split the dataset into a train set and a test set. We select 500 parameter-image pairs out of the 1000 pairs as the training set, in which $\lambda$ is from $\left [ 0.1,0.2 \right ) \cup \left [ 0.3,0.4 \right ) \cup \left [ 0.5,0.6 \right ) \cup \left [ 0.7,0.8 \right ) \cup \left [ 0.9,1.0 \right )$, and the rest are as the test set. Next, we train \Tech on the train set according to the normal method described in Section \ref{Methodology}, and evaluate our model on the test set.

\begin{figure*}[t]
    \centering
    \includegraphics[scale=0.38]{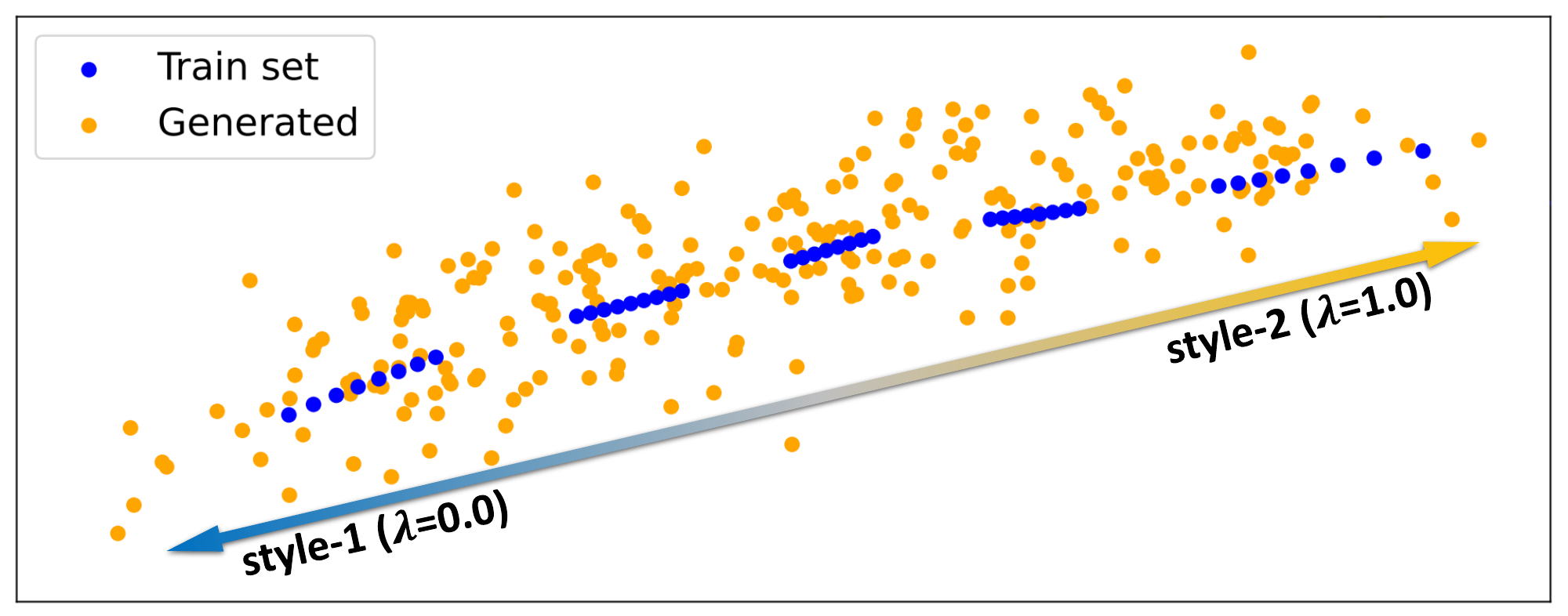}
    \caption{PCA in the latent space of the LoRA parameters of train set and generated by \Tech}
    \label{fig:appendix-cv-fig-3}
\end{figure*}

The results are shown in Figure \ref{fig:appendix-cv-fig-2}, where the input conditions are images chosen from the test set, which our model has never seen before. We find that the model can still generate images in the relevant style, which shows our model's generalizability. In addition, we visualized the training set parameters and the parameters generated by our model in the latent space by PCA \cite{pcaprincipalcomponentsanalysis} in Figure \ref{fig:appendix-cv-fig-3}. The blue dots represent the data used for training, and the place where the blue line is disconnected is left for testing. The orange dots represent the parameters generated by \Tech, and we find that our model can fit the entire distribution instead of only parts of the train set, illustrating the generalizability of the model.

\end{document}